\documentclass[letterpaper, 10 pt, journal, twoside]{IEEEtran}

\IEEEoverridecommandlockouts

\usepackage[noadjust]{cite}
\usepackage{times}
\usepackage{epsfig}
\usepackage{graphicx}
\usepackage{caption}
\usepackage{subcaption}
\newsavebox{\largestimage}
\usepackage{amsmath}
\usepackage{amssymb}
\usepackage{color}
\usepackage{url}
\makeatletter
\let\NAT@parse\undefined
\makeatother
\usepackage[colorlinks]{hyperref}
\usepackage{cleveref}
\usepackage{balance}
\usepackage{xparse}

\NewDocumentCommand\bbm{}{ \begin{bmatrix} }
\NewDocumentCommand\ebm{}{ \end{bmatrix} }

\NewDocumentCommand\Real{}{ \mathbb{R} }

\DeclareMathOperator*{\argmin}{arg\,min}

\newcommand{\irostitle}{Heteroscedastic Uncertainty for Robust Generative Latent Dynamics}

\title{Heteroscedastic Uncertainty for Robust\protect\\ Generative Latent Dynamics}
\author{Oliver Limoyo$^{1}$, Bryan Chan$^{1}$, Filip Mari\'{c}$^{1,2}$, Brandon Wagstaff$^{1}$, A. Rupam Mahmood$^{3}$ and Jonathan Kelly$^{1}$%
\thanks{Manuscript received: February 25, 2020; Revised June 14, 2020; Accepted July 31, 2020.}
\thanks{This paper was recommended for publication by Editor Dongheui Lee upon evaluation of the Associate Editor and Reviewers' comments.} 
\thanks{$^{1}$Authors are with the Space \& Terrestrial Autonomous Robotic Systems (STARS) Laboratory at the University of Toronto Institute for Aerospace Studies (UTIAS), Toronto, Ontario, Canada, M3H~5T6. Email: \texttt{<first name>.<last name>@robotics.utias.utoronto.ca}}
\thanks{$^{2}$ F. Mari\'{c} is jointly with the Laboratory for Autonomous Systems and Mobile Robotics (LAMOR) at the University of Zagreb, Croatia.}
\thanks{$^{3}$ A. R. Mahmood, Canada CIFAR AI Chair, Amii, is with the Reinforcement Learning and Artificial Intelligence Laboratory (RLAI) at the University of Alberta, Canada.}
\thanks{Digital Object Identifier (DOI): see top of this page.}
}

\begin{document}
\maketitle

\markboth{IEEE Robotics and Automation Letters. Preprint Version. Accepted July, 2020}
{Limoyo \MakeLowercase{\textit{et al.}}: \irostitle} 

\begin{abstract}
  Learning or identifying dynamics from a sequence of high-dimensional observations is a difficult challenge in many domains, including reinforcement learning and control. The problem has recently been studied from a generative perspective through \emph{latent dynamics}: high-dimensional observations are embedded into a lower-dimensional space in which the dynamics can be learned. Despite some successes, latent dynamics models have not yet been applied to real-world robotic systems where learned representations must be robust to a variety of perceptual confounds and noise sources not seen during training. In this paper, we present a method to jointly learn a latent state representation and the associated dynamics that is amenable for long-term planning and closed-loop control under perceptually difficult conditions. As our main contribution, we describe how our representation is able to capture a notion of heteroscedastic or input-specific uncertainty at test time by detecting novel or out-of-distribution (OOD) inputs. We present results from prediction and control experiments on two image-based tasks: a simulated pendulum balancing task and a real-world robotic manipulator reaching task. We demonstrate that our model produces significantly more accurate predictions and exhibits improved control performance, compared to a model that assumes homoscedastic uncertainty only, in the presence of varying degrees of input degradation.
\end{abstract}

\begin{IEEEkeywords}
  Model Learning for Control, Visual Learning, Reinforcement Learning, Calibration and Identification
\end{IEEEkeywords}

\section{Introduction}
\IEEEPARstart{I}{n many} robotics applications, the task-relevant state, such as position or velocity, cannot be directly measured and is only partially observed. Instead, the hidden or latent state must be recovered from high-dimensional sensory inputs. Similarly, the accompanying dynamics of the state are typically not available a priori. For complex systems, hand-engineered models may fail to properly represent the state or to adequately capture the dynamics.

A common family of solutions has been proposed recently in both the model-based reinforcement learning and sequence modelling literature---these approaches leverage the rich representative capacity of learned models and their ability to find a lower-dimensional representation of the input data and the dynamics. Despite some successes, however, existing methods have not yet been adapted for real-world robotic systems performing safety-critical tasks in environments where unexpected conditions (e.g., sensor noise, lighting variations, and unexpected obstructions) may occur.

\begin{figure}
	\centering
	\includegraphics[width=0.98\columnwidth]{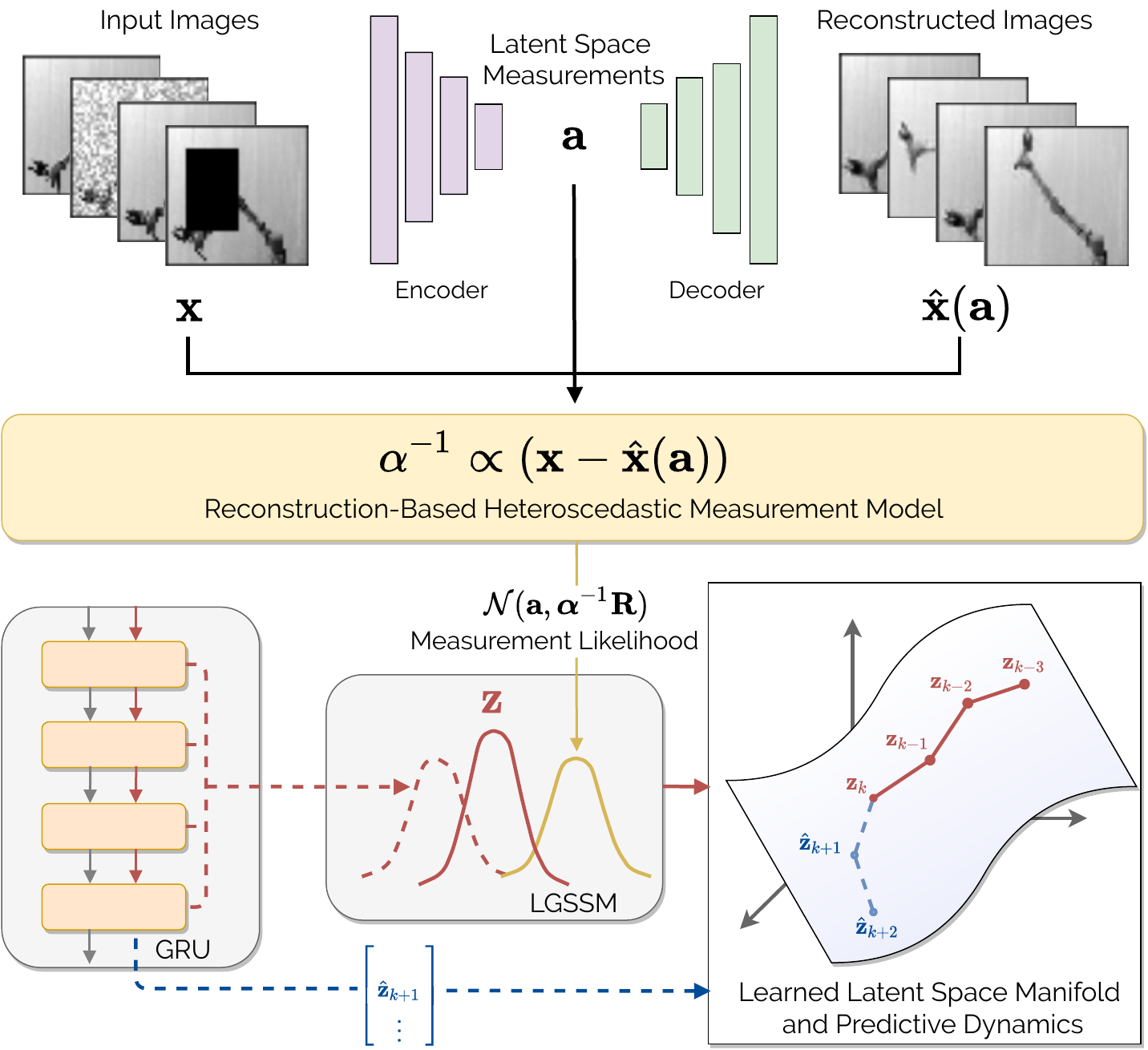}
	\caption{We learn a generative latent dynamics model that is robust to novel or out-of-distribution data at test time. To do so, we leverage the structure of a linear Gaussian state space model (LGSSM) with explicit uncertainty terms (e.g., the measurement uncertainty, $\mathbf{R}$), which can be set on a per-input basis depending on the approximated novelty of the data. 
	\label{fig:system_diagram}
}
\vspace{-4pt}
\end{figure}

In this letter, we propose a self-supervised approach (Fig. \ref{fig:system_diagram}) to jointly learn a probabilistic state representation and the associated dynamics, amenable for long-term planning and closed-loop control under perceptually difficult conditions. We take inspiration from recent papers that show how the structure of a linear Gaussian state space model (LGSSM) can be composed with one or more neural networks \cite{Karl2016-ok, Fraccaro2017-az, Chiappa2019-hy, Johnson2016-yr}. We are able to mitigate the potentially detrimental effects of observations that are far from the training set distribution by inflating the measurement uncertainty based on the estimated novelty of each observation. Our main contributions are:
	\begin{enumerate}
	\item a self-supervised method for learning a latent state representation and the associated dynamics while capturing a notion of heteroscedastic uncertainty at test time via novelty detection,
	\item a demonstration of image-based prediction and closed-loop control on a real-world robotic system using the learned latent dynamics that is robust to input degradations, and 
	\item an open source implementation of our algorithm.\footnote{See \url{github.com/utiasSTARS/robust-latent-srl} for code and supplementary material.}
	\end{enumerate}

\section{Related Work}

The literature on learned dynamics \cite{Ghahramani2000-ub, deisenroth2011pilco} and system identification \cite{Ljung1998-iv} is vast and diverse. Bearing in mind our application, we focus our review on prior and proximal work related to learning dynamics from image sequences; we view the problem through the lens of deep generative models or, more broadly, self-supervised representation learning of sequential data. 

 A significant portion of the existing sequential modelling literature has built upon either the autoencoder framework or its variational (VAE) alternative \cite{Kingma2013-kw, Rezende2014-ug}. The authors of \cite{Wahlstrom2015-wd} use the bottleneck of an autoencoder as the state of a deep dynamical model parameterized by a feedforward neural network. The model is trained with a reconstruction loss based on a one-step prediction. In \cite{Watter2015-ge} and \cite{Banijamali2017-nx}, a VAE is used to find both a low-dimensional latent representation of images and a transition model between consecutive image pairs. Extensions to the VAE framework for longer sequences are investigated in \cite{Gregor2015-xp} and \cite{Chung2015-dc} with a focus on recurrent neural network (RNN) architectures. Other lines of work combine structured and interpretable probabilistic graphical models with the flexibility of neural networks. In \cite{Haarnoja2016-kn}, an extended Kalman filter is employed in combination with a convolutional neural network (CNN) to output feature measurements and associated uncertainties. The Deep Markov Model, introduced in \cite{Krishnan2016-wz}, parametrizes the transition and measurement functions of a state-space model (SSM) with neural networks. Deep variational Bayes filters (DVBF) \cite{Karl2016-ok} reparametrize the variational inference problem to enable a recognition network to output transition parameters. In this way, reconstruction errors can backpropagate through the transitions directly. Other research \cite{Johnson2016-yr, Fraccaro2017-az} combines fast and exact inference subroutines with probabilistic graphical models and deep learning components. 
	
Learning dynamics from rich, high-dimensional data such as images has also been investigated in the context of model-based reinforcement learning. Such models have been used for both online planning \cite{Zhang2018-kt, Hafner2019-jj} and for generating synthetic trajectories for model-free policy learning \cite{Ha2018-ho, Sutton2012-ur}. These approaches emphasize learning forward transitions through a combination of reconstruction and additional auxiliary losses (e.g., reward prediction).

Our work is related to the field of novelty detection \cite{bishop1994novelty} and uncertainty quantification with Bayesian neural networks \cite{mackay1995probable}. Both of these fields address the general problem of evaluating the reliability of neural network outputs.
Existing Bayesian neural network methods use Monte Carlo dropout \cite{gal2015dropout} or ensembles \cite{lakshminarayanan2017simple} to capture the variance within the training data. Bayesian methods have not yet been shown to be able to capture accurate variances for regions far from the training distribution, however. Additionally, these networks often require a significant number of evaluations, which may compromise runtime efficiency on physical platforms. Taking into account these limitations, we aim to improve the reliability of network outputs by leveraging techniques from the field of novelty detection. Relevant prior work includes \cite{Richter2017-db}, where an autoencoder forms part of an image-based collision prediction system for a ground vehicle. When the autoencoder reconstruction loss is large, the input images are considered to be OOD and the system reverts to a more conservative (safer) collision prediction algorithm.
In \cite{pomerleau1993input} and \cite{Amini2018VariationalAF}, similar reconstruction-based approaches to novelty detection are presented for end-to-end learning of a control policy for autonomous driving. We also use a reconstruction-based approach to novelty detection but adapt it to the context of latent dynamics.
	
In this paper, we build on learning methods for sequential data that take advantage of the structure of a SSM \cite{Karl2016-ok, Fraccaro2017-az, Chiappa2019-hy, Johnson2016-yr}. However, our work differs from and extends these approaches in the following ways: 1) we combine ideas from the novelty detection literature to substantially enhance robustness to input degradations not seen during training, by making use of the model's generative capabilities and probabilistic representation, and 2) we provide experimental results from a real-world robotic prediction and control task, demonstrating that our approach can be successfully applied to physical systems.

\section{Methodology}
\label{methods}

We assume that our training set is uncorrupted and contains the dynamics of interest. In Section \ref{generative}, we first consider the problem of modelling the dynamics of a sequence of $K$ high-dimensional observations, $\{\mathbf{x}_{k}\}_{k=1}^{K}$, generated by a random exploration policy. At test time, we require our system to be able to robustly predict future states under various difficult perceptual conditions not seen during training. Neural networks are known to generalize poorly when queried with novel or OOD inputs \cite{bishop1994novelty}. 
For real-world, safety-critical applications, this failure to generalize is a key limiting factor. In Section \ref{heteroscedastic}, we demonstrate how a notion of heteroscedastic uncertainty can be introduced in order to mitigate this issue. In Section \ref{mpc}, we define a model predictive control (MPC) problem to demonstrate a use case of our learned robust image dynamics.
Finally, in Section \ref{network}, we provide details on our specific network architecture and training procedure.

\subsection{Generative Latent Dynamics}
\label{generative}

The task of identifying latent dynamics can be formulated as one of learning a lower-dimensional latent state representation $\{\mathbf{z}_{k}\}_{k=1}^{K},\, \mathbf{z}_{k} \in \Real^{n}$ and the respective dynamics or state transition function $\mathbf{z}_{k+1} \sim p(\mathbf{z}_{k+1} |\,\mathbf{z}_{k}, \mathbf{u}_{k})$, with control input or action $\mathbf{u}_{k} \in \Real^{m}$. For our specific application, we consider the case in which the input measurements $\mathbf{x}_{k} \in \Real^{w \times h}\!$ are images captured by a camera of a scene that contains the underlying dynamics. In order to model the dynamics, we employ a LGSSM as follows,
	\begin{equation} \label{eq:transition}
	p(\mathbf{z}_{k+1} |\,\mathbf{z}_{k}, \mathbf{u}_{k}) = \mathcal{N}(\mathbf{A}_{k}\mathbf{z}_{k} + \mathbf{B}_{k}\mathbf{u}_{k}, \mathbf{Q}),
	\end{equation}
	with local transition matrix $\mathbf{A}_{k} \in \Real^{n \times n}$, local control matrix $\mathbf{B}_{k} \in \Real^{n \times m}$, and constant process noise covariance $\mathbf{Q} \in \Real^{n \times n}$. The accompanying measurement model is 
	\begin{equation} \label{eq:measurement}
	p(\mathbf{a}_{k} |\,\mathbf{z}_{k}) = \mathcal{N}(\mathbf{C}_{k}\mathbf{z}_{k}, \mathbf{R}),
	\end{equation}
	where measurement $\mathbf{a}_{k} \in \Real^{l} \sim p(\mathbf{a}_{k} |\,\mathbf{x}_{k})$ is extracted from image $\mathbf{x}_{k}$ as in \cite{Fraccaro2017-az}, $\mathbf{C}_{k} \in \Real^{l \times n}$ is the local measurement matrix, and $\mathbf{R} \in \Real^{l \times l}$ is the measurement noise covariance. We choose to add the structure of the LGSSM to be able to explicitly reason about uncertainties in an interpretable manner. The joint probability distribution of a sequence of measurements $\mathbf{a} = \{\mathbf{a}_{k}\}_{k=1}^{K}$ and latent states $\mathbf{z} = \{\mathbf{z}_{k}\}_{k=1}^{K}$ can then be defined as
	\begin{equation} \label{eq:joint_probability}
	\begin{split}
	p(\mathbf{a}, \mathbf{z}\,|\,\mathbf{u}) & = p(\mathbf{a}\,|\,\mathbf{z})\,p(\mathbf{z} \,|\,\mathbf{u})  \\
	& = \prod_{k=1}^{K} p(\mathbf{a}_{k} |\,\mathbf{z}_{k}) \cdot p(\mathbf{z}_{1}) \prod_{k=1}^{K - 1} p(\mathbf{z}_{k + 1} |\, \mathbf{z}_{k}, \mathbf{u}_{k}),
	\end{split}
	\end{equation}
	where $\mathbf{z}_{1} \in \Real^{n} \sim p(\mathbf{z}_{1})$ is the initial state from a given fixed distribution $p(\mathbf{z}_{1}) = \mathcal{N}(\mathbf{0}, \mathbf{\Sigma}_{1})$. Note that, as a byproduct of employing a LGSSM, exact inference can be carried out for the posterior $p(\mathbf{z}\,|\,\mathbf{a}, \mathbf{u})$ using the common Kalman filter (KF) \cite{kf1960} or Rauch-Tung-Striebel (RTS) smoothing equations \cite{Rauch1965-pc}.
	
Following the insights of \cite{Fraccaro2017-az} and \cite{Chiappa2019-hy}, we pose the problem of approximating the intractable posterior of the latent measurements given the images, $p(\mathbf{a}\,|\,\mathbf{x})$, in terms of variational inference. As is commonly done in the variational inference literature, we use a recognition network to output the mean and standard deviation of the (assumed) Gaussian posterior,
	\begin{equation} \label{eq:VAE_q}
	q_{\phi}(\mathbf{a}\,|\,\mathbf{x}) = \prod_{k=1}^{K} q_{\phi}(\mathbf{a}_{k} |\,\mathbf{x}_{k}).
	\end{equation}
 Our network is comprised of a CNN followed by a fully connected layer, based on the architecture in \cite{Ha2018-ho}. We use transposed convolution layers in the generative decoder to output the parameters of the assumed Gaussian or Bernoulli conditional distribution,
	\begin{equation} \label{eq:VAE_p}
	p_{\theta}(\mathbf{x}\,|\,\mathbf{a}) = \prod_{k=1}^{K} p_{\theta}(\mathbf{x}_{k} | \,\mathbf{a}_{k}).
	\end{equation}
  A Gated Recurrent Unit (GRU) network \cite{cho2014properties} is trained to regress the matrices $\mathbf{A}_{k}, \mathbf{B}_{k}, \mathbf{C}_{k}$,
	\begin{equation} \label{eq:dynamics}
	\mathbf{A}_{k}, \mathbf{B}_{k}, \mathbf{C}_{k}, \textbf{h}_{k+1} = g_{\psi}(\textbf{z}_{k}, \textbf{u}_{k},  \textbf{h}_{k}),
	\end{equation}
	where $\textbf{h}_{k} \in \Real^{v}$ is the recurrent hidden state at time step $k$. The GRU network outputs a linear combination of base matrices that are independent of the input, and are learned globally (see \cite{Karl2016-ok, Fraccaro2017-az} for a more detailed exposition). In order to learn or identify the parameters $\{\phi, \theta, \psi\}$, we use the reparameterization trick \cite{Kingma2013-kw, Rezende2014-ug} to maximize a lower bound \cite{Jordan1999-lp} of the marginal log-likelihood of the data, $p(\mathbf{x}\,|\,\mathbf{u}) = \int p(\mathbf{x}, \mathbf{a}, \mathbf{z}\, |\, \mathbf{u})\,d\mathbf{a}\,d\mathbf{z}$ via stochastic gradient updates by sampling from
	\begin{equation} \label{eq:elbo}
	\ln p(\mathbf{x}\,|\,\mathbf{u}) \geq \mathbb{E}_{q_{\phi}(\mathbf{a}\,|\,\mathbf{x})}[\ln p_{\theta}(\mathbf{x}\,|\,\mathbf{a}) - \text{KL}(q_{\phi}(\mathbf{a}\,| \,\mathbf{x}) \| p(\mathbf{a}\,|\,\mathbf{u}))],
	\end{equation}
	where
	\begin{equation} \label{eq:prior}
	p(\mathbf{a}\,|\,\mathbf{u}) 
	= \frac{p(\mathbf{a}, \mathbf{z}\,|\,\mathbf{u})}{p(\mathbf{z}\,|\,\mathbf{a}, \mathbf{u})}.
	\end{equation}
We refer readers to \cite{Fraccaro2017-az} for a derivation of the lower bound used in this work.
	
\subsection{Heteroscedastic Uncertainty via Novelty Detection}
\label{heteroscedastic}

\begin{figure}
\centering
\includegraphics[width=0.99\columnwidth]{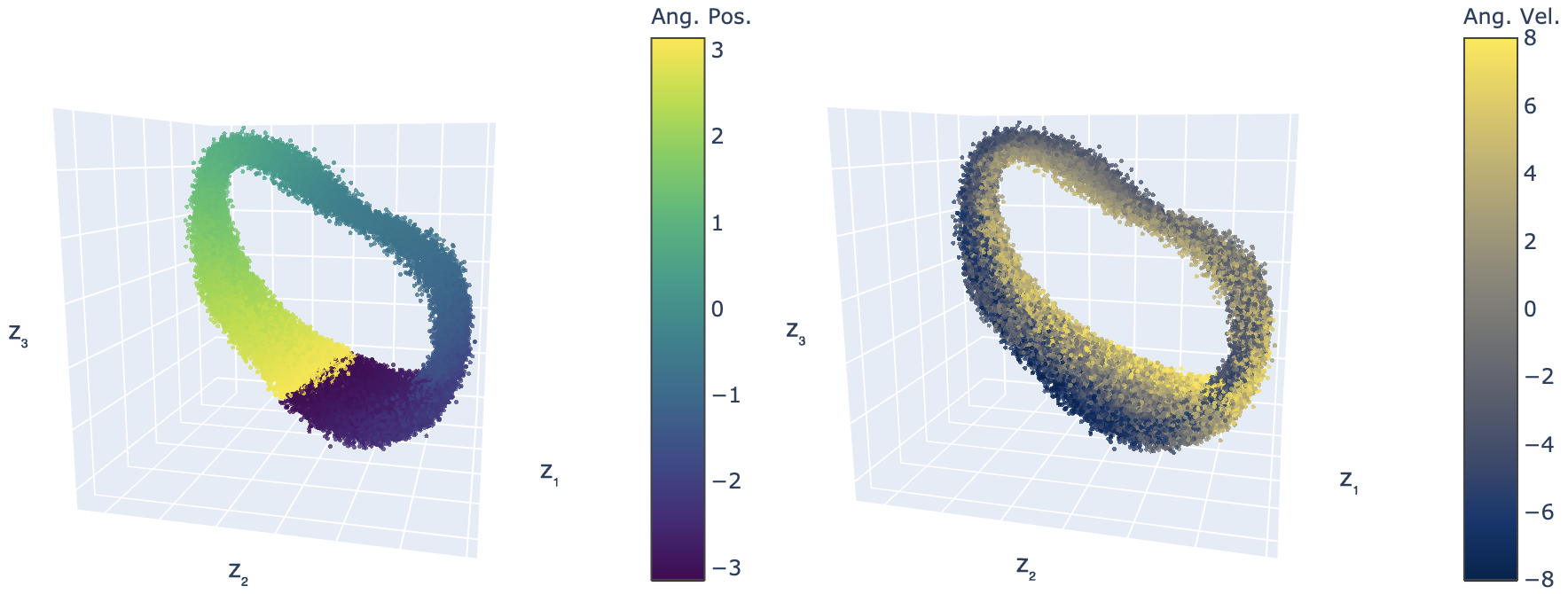}
\caption{Smoothed, inferred latent space, $\mathbf{z}_{k} \in \Real^{3}$, for the pendulum task. \emph{Left:} Points are colour-coded by ground truth joint angle. \emph{Right:} Points are colour-coded by angular velocity.}
\label{fig:latent_state}
\end{figure}
The LGSSM structure allows us to set a per time step uncertainty, $\mathbf{R}_{k}$, instead of a constant uncertainty, $\mathbf{R}$ \cite{sinopoli2004kalman}. Crucially, in (\ref{eq:dynamics}), we explicitly learn the parameters $\mathbf{A}_{k}, \mathbf{B}_{k}, \mathbf{C}_{k}$ based on past latent states $\{\mathbf{z}_{i}\}_{i=1}^{k}$, as opposed to past measurements (features) $\{\mathbf{a}_{i}\}_{i=1}^{k}$ as done in \cite{Fraccaro2017-az} and \cite{Chiappa2019-hy}. We do this to decouple the dynamics from the measurements. If required, this allows our model to prioritize the dynamics and to reduce the influence of the measurement $\textbf{a}_{k}$ on the state $\textbf{z}_{k}$ during inference by inflating the respective uncertainty $\mathbf{R}_{k}$. We note that $\mathbf{R}_{k}$ can be constrained to be diagonal without loss of generality when our latent states have no predefined meaning \cite{Murphy2012-uk}, as in this case. By doing so, we reduce the number of free parameters and improve numerical stability. 

	To account for OOD data, we raise our density to a power $\alpha_{k}(\mathbf{a}_{k}, \mathbf{x}_{k}) \in \Real^{+}$,
	\begin{equation} \label{eq:alpha}
	p^{\alpha_{k}} (\mathbf{a}_{k} |\,\mathbf{z}_{k}).
	\end{equation}
	Raising a density to a power is proposed as a technique in \cite{cao2014generalized} to modify a product of experts such that each expert's reliability can be scaled based on the input datapoint. The Kalman filter measurement update can be interpreted as a product of experts, as shown by (\ref{eq:joint_probability}). Raising a density to a power has also been used in Markov chain Monte Carlo (MCMC) algorithms to anneal distributions and to balance probabilistic models with varying degrees of freedom \cite{urtasun20063d}. In our case, where the original distribution is a Gaussian, the resulting distribution remains Gaussian and the operation can be shown to be equivalent to scaling the precision matrix (or the measurement covariance),
	\begin{equation} \label{eq:equivalence}
	p^{\alpha_{k}} (\mathbf{a}_{k} |\,\mathbf{z}_{k}) = \frac{1}{Z} \exp{(-\frac{1}{2}\mathbf{r}_{k}^{T}\alpha_{k}\mathbf{R}^{-1}\mathbf{r}_{k})},
	\end{equation}
	where $\mathbf{r}_{k} = (\mathbf{a}_{k} - \mathbf{C}_{k}\mathbf{z}_{k})$ is the innovation or residual and $\mathbf{R}^{-1}_{k} = \alpha_{k}\mathbf{R}^{-1}$ is the reweighed precision matrix. Based on our network's generative capabilities, we use the following reconstruction measure as a proxy for the network's confidence or reliability,
	\begin{equation} \label{eq:measure}
	\alpha_{k}(\mathbf{a}_{k}, \mathbf{x}_{k}) =  \log{(1 + \frac{\bar{L}_{\text{train}}} {L_{k}})},
	\end{equation}
	where $\bar{L}_{\text{train}} = \frac{1}{N}\sum_{n=1}^{N} \| \mathbf{x}_{n} - \hat{\mathbf{x}}_{n}(\mathbf{a}_{n}) \|_{F}^{2}$ is the average reconstruction loss over the $N$ images in our training set and  $L_{k} = \| \mathbf{x}_{k} - \hat{\mathbf{x}}_{k}(\mathbf{a}_{k}) \|_{F}^{2}$ is the reconstruction loss for image $\mathbf{x}_{k}$. The reconstructed images $\hat{\mathbf{x}}$ are produced by the generative decoder defined by (\ref{eq:VAE_p}); an image with a larger reconstruction error produces a smaller measurement precision (or a larger measurement covariance). Similar approaches to reconstruction-based novelty detection have been used in \cite{pomerleau1993input}, \cite{Richter2017-db} and \cite{Amini2018VariationalAF}.

\subsection{Model Predictive Control in Latent Space} 
\label{mpc}

As a control problem, we consider the task of producing a specified goal image from a given initial image (i.e., mapping from pixels to pixels); we can use a history of multiple images if needed. We solve for the optimal controls over a prediction horizon of length $T$ by way of MPC with the learned latent dynamics. We first embed a window of $T_{i}$ initial images $\{\mathbf{x}^{i}_{k}\}_{k=1}^{T_{i}}$ into measurements $\{\mathbf{a}^{i}_{k}\}_{k=1}^{T_{i}}$ using the encoder network (\ref{eq:VAE_q}). From both $\{\mathbf{x}^{i}_{k}\}_{k=1}^{T_{i}}$ and $\{\mathbf{a}^{i}_{k}\}_{k=1}^{T_{i}}$, we can calculate the heteroscedastic weighing factors $\{\alpha^{i}_{k}\}_{k=1}^{T_{i}}$ using (\ref{eq:measure}) and rescale the measurement precision matrices $\{(\mathbf{R}^{i}_{k})^{-1}\}_{k=1}^{T_{i}} = \{\alpha^{i}_{k}\mathbf{R}^{-1}\}_{k=1}^{T_{i}}$ accordingly. We can then solve for a window of initial latent states $\{\mathbf{z}^{i}_{k}\}_{k=1}^{T_{i}}$ by exact inference with the KF or RTS smoothing equations and the learned dynamics from (\ref{eq:dynamics}), given the control inputs $\{\mathbf{u}^{i}_{k}\}_{k=1}^{T_{i}}$. In a similar manner, we can solve for a window of goal latent states $\{\mathbf{z}^{g}_{k}\}_{k=1}^{T_{g}}$ for a given window of $T_{g}$ goal images $\{\mathbf{x}^{g}_{k}\}_{k=1}^{T_{g}}$ (potentially, a single image repeated $T_{g}$ times) and control inputs $\{\mathbf{u}^{g}_{k}\}_{k=1}^{T_{g}} = \mathbf{0}$.  

We define the optimization problem for the controls $\mathbf{u}^{*}$ over the next $T$ time-steps as
	\begin{equation} \label{eq:mpc}
	\begin{split}
	\mathbf{u}^{*} = \argmin_{\mathbf{u}} & \sum_{k=1}^{T} (\mathbf{z}_{k+1} - \mathbf{z}^{g}_{T_{g}})^{T}\mathbf{Q}_{\text{mpc}}(\mathbf{z}_{k+1} - \mathbf{z}^{g}_{T_{g}}) \\ &\qquad + \mathbf{u}_{k}^{T} \mathbf{R}_{\text{mpc}} \mathbf{u}_{k},\\
	\textrm{s.t.} \quad & \mathbf{z}_{k+1} = \mathbf{A}_{k} \mathbf{z}_{k} + \mathbf{B}_{k} \mathbf{u}_{k}, \\
		& \mathbf{z}_{1} = \mathbf{z}^{i}_{T_{i}},
	\end{split}
	\end{equation}
where $\mathbf{u} = \{\mathbf{u}_{1}, ..., \mathbf{u}_{T}\}$, $\mathbf{Q}_{\text{mpc}} \in \Real^{n \times n}$ is positive semidefinite, $\mathbf{R}_{\text{mpc}} \in \Real^{m \times m}$ is positive definite, and $\mathbf{z}^{i}_{T_{i}}$ is set as the initial latent state and $\mathbf{z}^{g}_{T_{g}}$ as the goal latent state. Transition matrices $\mathbf{A}_{1:T}$ and control matrices $\mathbf{B}_{1:T}$ output from the GRU network (\ref{eq:dynamics}) given an initial guess for $\textbf{u}$. We can then minimize (\ref{eq:mpc}) for $\textbf{u}^{*}$ with any common convex optimization technique; we use the \texttt{CVXPY} modelling language \cite{diamond2016cvxpy}. As commonly done in linear MPC, we use the updated solution to produce a new set of transition and control matrices and iterate until convergence. We send the first control input $\textbf{u}^{*}_{1}$ to the system and then observe the resulting image $\mathbf{x}_{2}$, encoding it into $\mathbf{a}_{2}$. This process continues with the controls from the previous horizon as the new initial guess.

\subsection{Training Procedure}
\label{network}

Our cost function is the negative of the lower bound of the marginal log-likelihood of the data, as shown by (\ref{eq:elbo}). Given a sequence of input images of length $I$ from our training set, $\mathbf{x} = \{\mathbf{x}_{i}\}_{i=1}^{I}$, we approximate the expectation by taking a single posterior sample $\tilde{\mathbf{a}} \sim q_{\phi}(\mathbf{a}\,|\,\mathbf{x})$ of the encoded images. We can then use $\tilde{\mathbf{a}}$ within the LGSSM and analytically obtain $p(\mathbf{z}\,|\,\mathbf{a}, \mathbf{u})$ from: 1) an assumed initial distribution $p(\mathbf{z}_{1})$, 2) the KF or RTS smoothing equations, 3) the locally linear dynamics and measurement models from the GRU network, and 4) the known control inputs $\mathbf{u} = \{\mathbf{u}_{i}\}_{i=1}^{I}$. Next, we  sample from $\tilde{\mathbf{z}} \sim p(\mathbf{z}\,|\,\mathbf{a}, \mathbf{u})$ and analytically obtain $p(\mathbf{a}, \mathbf{z}\,|\,\mathbf{u})$, as shown in (\ref{eq:joint_probability}), by way of (\ref{eq:transition}) and (\ref{eq:measurement}). Lastly, $p_{\theta}(\mathbf{x}\,|\,\mathbf{a})$ is assumed to be a Gaussian or Bernoulli distribution with parameters from the decoder network, as is commonly done with VAEs. Given all of the distributions, we can calculate the likelihood or cost of the samples $\tilde{\mathbf{a}}$ and $\tilde{\mathbf{z}}$ and use stochastic gradient descent to update the parameters $\{\phi, \theta, \psi\}$ of the respective encoder, decoder, and GRU networks. Detailed descriptions of the respective network architectures and the associated hyperparameters are available in our supplementary material.

\section{Experiments} \label{experiments}

\begin{figure}
	\centering
	\includegraphics[height=5.05cm]{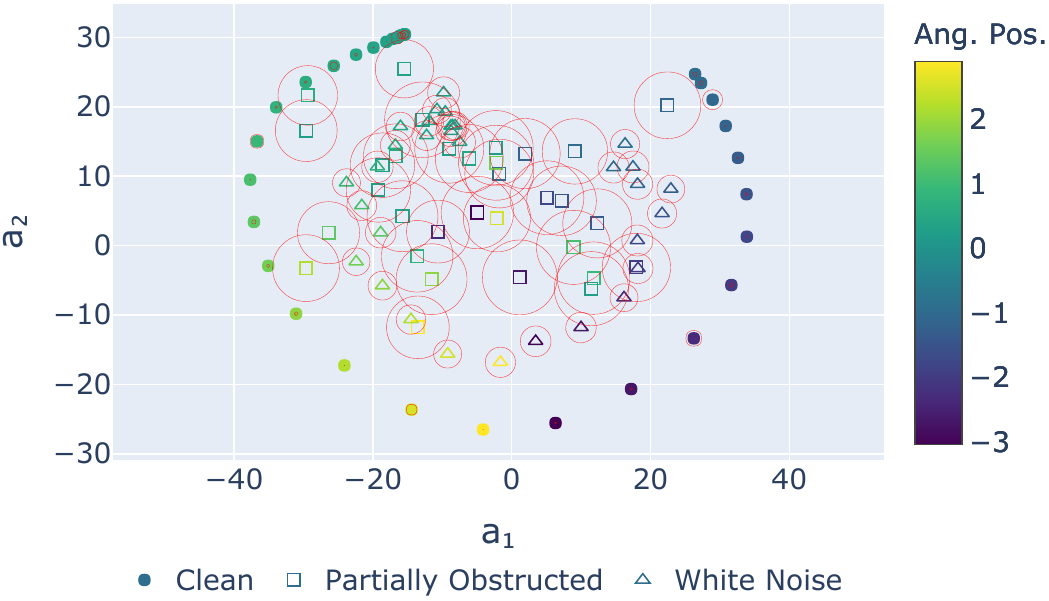}
	\caption{The extracted latent measurements for the pendulum task, $\mathbf{a}_{k} \in \Real^{2}$, from clean images (filled circles) and corrupted images (hollow squares and triangles) with their respective covariances (in red). Note that the covariances of the clean measurements are small and more difficult to visually distinguish.}
	\label{fig:latent_measurement}
\end{figure}
\begin{figure*}[t]
	\centering
    \begin{subfigure}[t]{0.99\textwidth}
        \includegraphics[width=\textwidth]{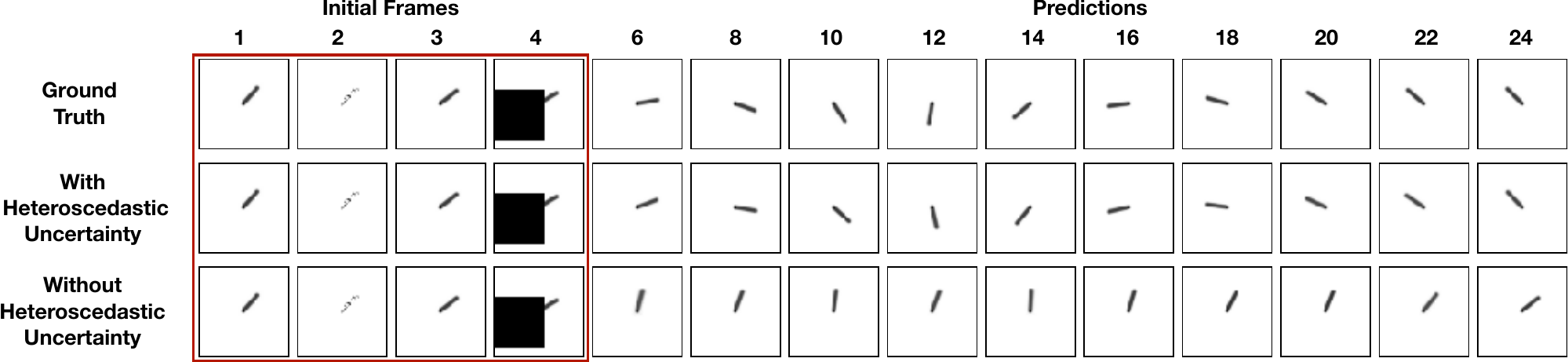}
        \caption{Simulated pendulum task: Images 2 and 4 are corrupted with noise and an obstruction, respectively.}
        \label{fig:pendulum_predict}
    \end{subfigure}
    \par\bigskip 
    \begin{subfigure}[t]{0.99\textwidth}
        \includegraphics[width=\textwidth]{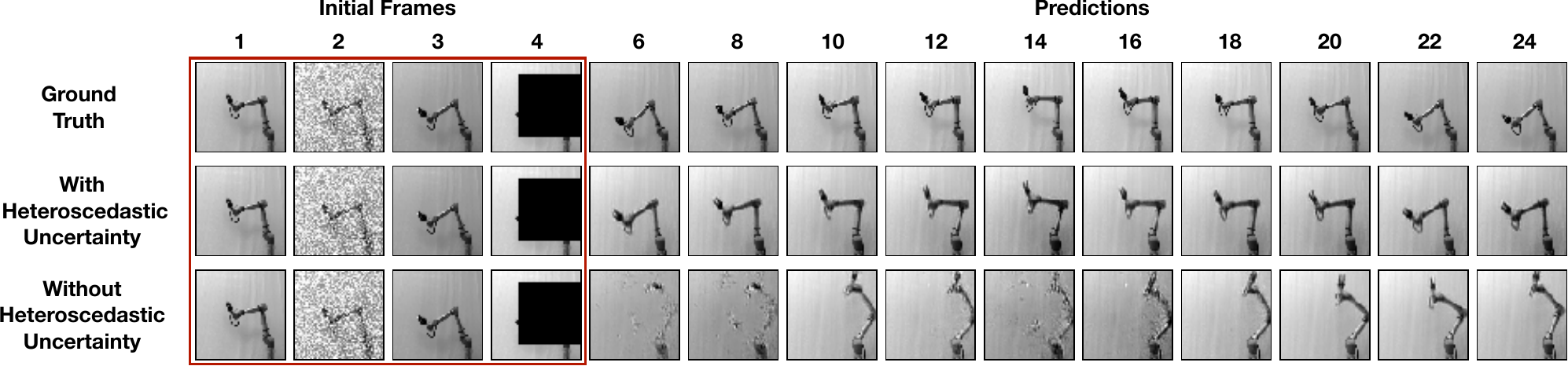}
        \caption{Real-world visual reaching task: Images 2 and 4 are corrupted with noise and an obstruction, respectively.}
        \label{fig:reacher_predict}
    \end{subfigure}
    \par\bigskip 
    \begin{subfigure}[t]{0.99\textwidth}
        \includegraphics[width=\textwidth]{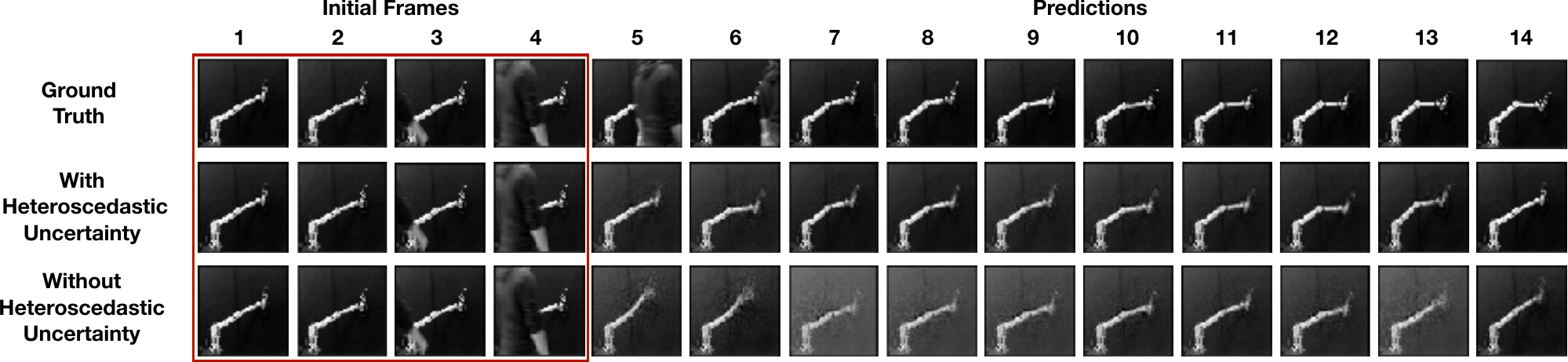}
        \caption{Real-world visual reaching task: Image 4 is corrupted with a person obstructing the camera.}
        \label{fig:reacher_predict_real_obs}
    \end{subfigure}
	\caption{Image prediction results. For each set, the first four columns, marked by a red rectangle, represent the frames used to initialize our state. The remaining images are generated from future latent state predictions by our learned dynamics model and the decoder network. The number at the top identifies the time step for the images in that column. \emph{Top row:} The ground truth images. \emph{Middle row:} Image predictions with use of heteroscedastic uncertainty. \emph{Bottom row:} Image predictions without use of heteroscedastic uncertainty.}
\end{figure*}
We evaluated our approach by conducting experiments involving two different image-based control tasks: a modified simulated pendulum task from OpenAI Gym \cite{brockman2016openai} and a real-world manipulator 2D reaching task. In both cases, we tested the predictive and control capability of our model under challenging conditions using image data alone. The high-dimensional observations (i.e., images with thousands of pixels) and the highly nonlinear mapping from the underlying system dynamics to the image make these tasks non-trivial; naive application of classical optimal control would be infeasible. We chose a random, zero-mean Gaussian (i.e., random control input) exploration policy to collect training trajectories with a control or action repeat of three \cite{mnih2013playing}.

\subsection{Simulated Visual Pendulum Task}

We began with a `toy' environment as an initial test to allow us to directly visualize and analyze what our model learned. As our only modification to the existing OpenAI environment, we used grayscale images as inputs, $\mathbf{x} \in \Real^{64 \times 64}$, instead of the pendulum joint angle and velocity; the control input $u \in \Real$ (i.e., the input torque to the pendulum) remained unchanged. We chose a latent space $\mathbf{z} \in \Real^{3}$ and a measurement space $\mathbf{a} \in \Real^{2}$. During training, we collected a total of $N = 2048$ trajectories of length 32 to train our model.

\subsubsection{Visualization Experiments}
\label{pend_vexp}

The trained model learned a latent state representation that resembles a ring-like manifold, as shown in Fig. \ref{fig:latent_state}. Each point in the latent space in  Fig. \ref{fig:latent_state} is colour-coded based on the ground truth angular position and velocity of the pendulum (on the left and right, respectively). Fig. \ref{fig:latent_measurement} visualizes the measurements, $\mathbf{a}_{k}$, and their respective covariances, $\mathbf{R}_{k}$, for three different input image conditions: clean, partially obstructed, and noisy (i.e., with added Gaussian pixel noise). The uncertainties of the measurements associated with the degraded images are substantially larger than the those of the clean images, as we would expect.

\subsubsection{Prediction Experiment}
\label{pend_pexp}
\begin{figure*}
    \centering
	\captionsetup{aboveskip=0pt}
    \begin{subfigure}[t]{0.28\textwidth}
        \includegraphics[width=\textwidth]{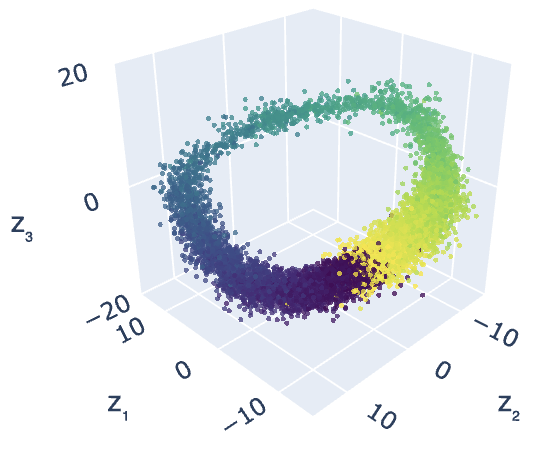}
        \caption{}
        \label{fig:robust_clean}
        \usebox{\largestimage}
    \end{subfigure}
    ~
    \begin{subfigure}[t]{0.28\textwidth}
        \includegraphics[width=\textwidth]{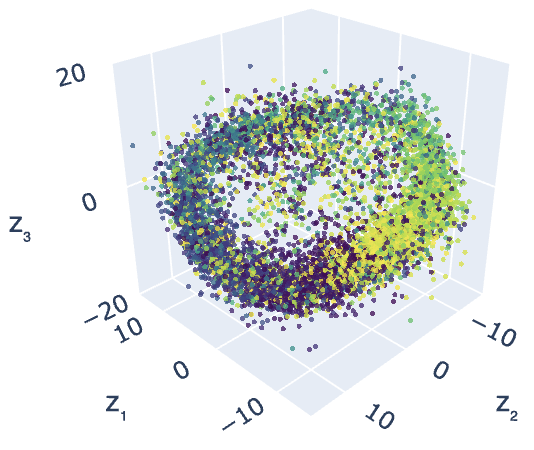}
        \caption{}
        \label{fig:robust_noisy_naive}
    \end{subfigure}
    ~
    \begin{subfigure}[t]{0.28\textwidth}
        \includegraphics[width=\textwidth]{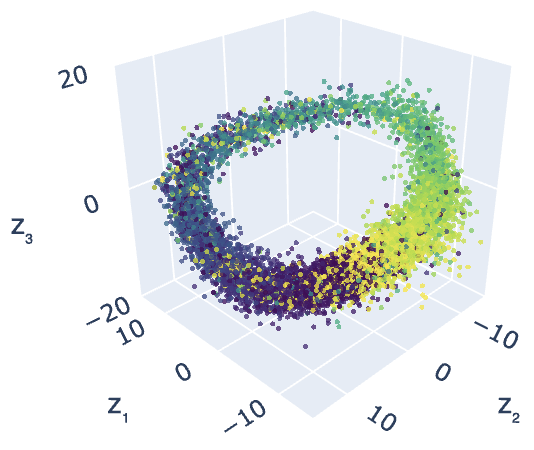}
        \caption{}
        \label{fig:robust_noisy_ood}
    \end{subfigure}
    ~~
    \begin{subfigure}[t]{0.07\textwidth}
        \raisebox{\dimexpr\ht\largestimage+0.125\height}{\includegraphics[width=\textwidth]{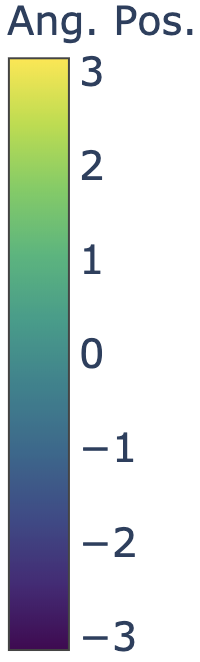}}
    \end{subfigure}
    \caption{Predicted or generated latent states, $\mathbf{\hat{z}}_{k} \in \Real^{3}$, for the pendulum task. The latent state is initialized as follows. \emph{Left}: 4 clean images, \emph{Middle}: 2 out of 4 initial images obstructed randomly and no heteroscedastic uncertainty, \emph{Right}: 2 out of 4 initial images obstructed randomly and with heteroscedastic uncertainty. Points are colour-coded by the ground truth joint angle.} 
    \label{fig:robust_structure}
\end{figure*}
We tested the capability of the model to predict 20 future latent states, $\hat{\mathbf{z}}_{5:24}$, based on just four initial images from a validation set, $\mathbf{x}_{1:4}$. To compute the future states, we used the same control inputs $\mathbf{u}_{4:23}$ which were sent to the pendulum and produced the ground-truth images $\mathbf{x}_{5:24}$. The initial images were first embedded into measurements $\mathbf{a}_{1:4}$ using the encoder network. We recovered the initial latent states $\mathbf{z}_{1:4}$ with the standard Kalman filter equations. We then `rolled-out' a trajectory, using the learned transition and control matrices from the GRU network, in order to arrive at the predicted future latent states $\hat{\mathbf{z}}_{5:24}$. The learned measurement matrices, also from the GRU network, were used to generate the predicted measurements $\hat{\mathbf{a}}_{5:24}$ from the predicted latent states $\hat{\mathbf{z}}_{5:24}$. Finally, the decoder network utilized the predicted measurements to generate predicted images $\hat{\mathbf{x}}_{5:24}$. The results are shown in Fig. \ref{fig:pendulum_predict}. Two of the initial images were corrupted, by noise (Image 2) and by a single-frame obstruction (Image 4). For the baseline model, with no notion of heteroscedastic uncertainty (Row 3), the added noise and obstruction in the initial images caused the future predictions to be highly inaccurate when compared to the ground-truth images (Row 1). On the other hand, when using our heteroscedastic uncertainty weighing (Row 2), the model was capable of effectively ignoring the corrupted images and predicting the future states accurately.

As a final experimental verification in the pendulum environment, we visualized sets of predicted future latent states (Fig. \ref{fig:robust_structure}). Each subplot in Fig. \ref{fig:robust_structure} shows the predicted latent states under specific conditions. On the left, we visualized the predicted latent states when all four of the initial images were unobstructed and noise-free. For the middle and right subfigures, we corrupted two of the four images with randomly placed obstructions. The middle subfigure demonstrates that, without heteroscedastic uncertainty modelling, the ring-like structure and the ordering of the points (in terms of their colour-coded joint angles) is compromised. In the right subfigure, we show that the use of heteroscedastic uncertainties better preserves the manifold structure and joint angle ordering despite the presence of the same obstructions.

\subsection{Real-World Visual Reacher}
\label{sec:vis_reacher}
\begin{figure}[b]
	\centering
	\includegraphics[width=0.70\columnwidth]{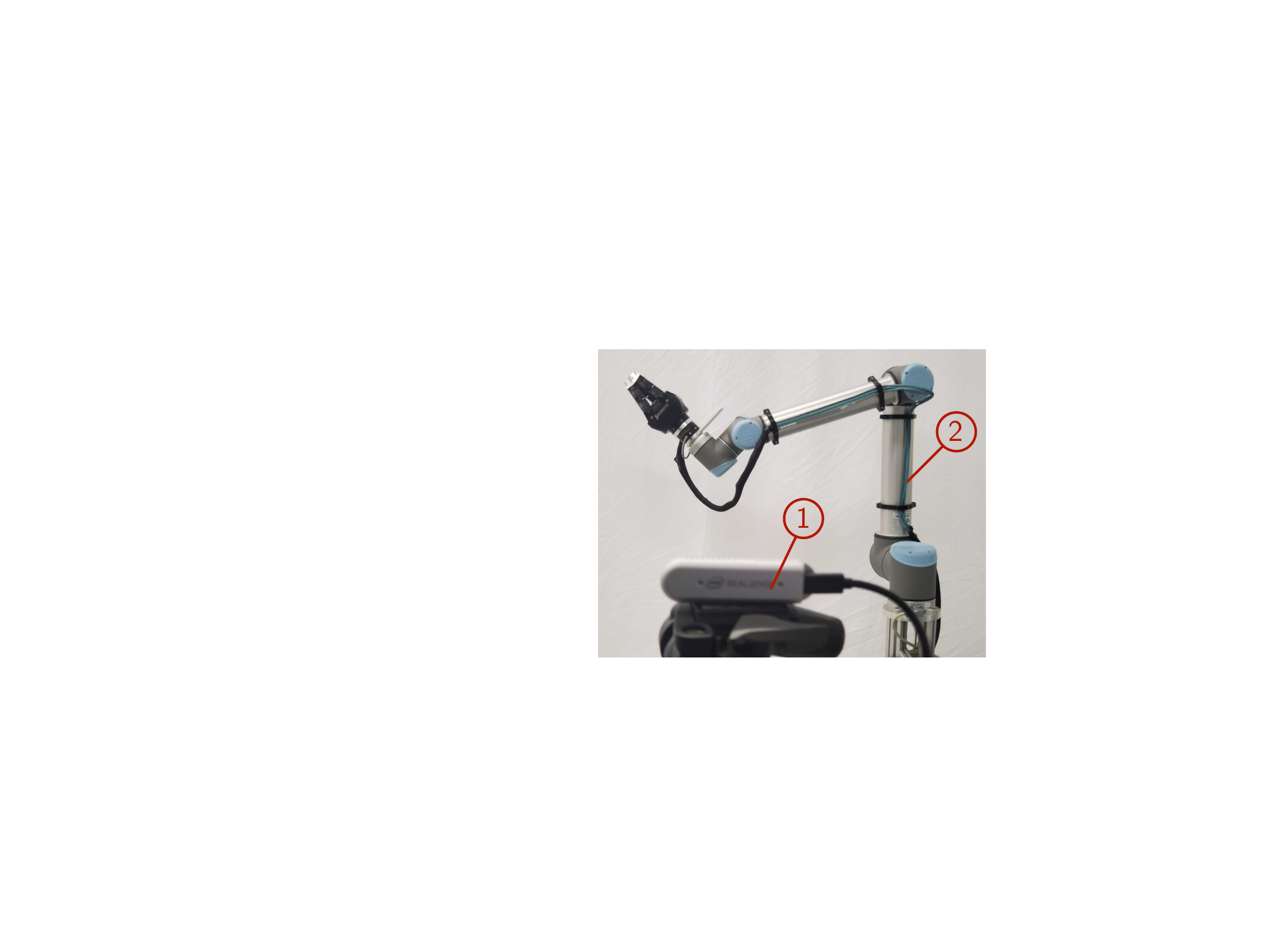}
	\caption{Experimental setup for our real-world visual reaching task: (1) UR10 manipulator and (2) Intel RealSense D435 camera.}
	\label{fig:experimental_setup}
\end{figure}
We tested our model on a more difficult (real-world) 2D visual reaching task. This is a continuous control task based on the real-world robot reinforcement learning benchmark `UR-Reacher-2' described in \cite{mahmood2018benchmarking}. We collected a dataset of images of a robot arm performing the reaching task, along with the corresponding control inputs. Our experimental setup is shown in Fig. \ref{fig:experimental_setup}. The control inputs $\mathbf{u} \in \Real^{2}$ were restricted to the velocities of joints 2 and 3 of the UR10 arm. Input images were captured by an external, fixed camera.
We emphasize that the joint encoder readings were not used to train our dynamics model; we kept track of the joint angles solely for ground truth evaluation and analysis. The input images were downsampled to $\mathbf{x} \in \Real^{64 \times 64}$ and converted to grayscale. We chose a latent space $\mathbf{z} \in \Real^{10}$ and a measurement space $\mathbf{a} \in \Real^{4}$. In order to enforce synchronized communication with reduced control latency, we relied on the SenseAct framework \cite{mahmood2018benchmarking} to gather our data. We collected a total of $N=1024$ trajectories for training, each with a length of 15 time steps, where each step had a duration of 0.5 seconds.

\subsubsection{Prediction Experiment}

As shown in Fig. \ref{fig:reacher_predict}, we tested our model's predictive capability in the presence of noise (Image 2) and a single-frame obstruction (Image 4). We followed the same procedure detailed in Section \ref{pend_pexp}. Similar to the results for the pendulum experiment, the model was better able to predict future states in the presence of corrupted measurements when it incorporated heteroscedastic uncertainty. Fig. \ref{fig:reacher_predict_real_obs} demonstrates the robustness of the model to a real (non-simulated) obstruction (i.e., a person blocking the camera) in Image 3.
 
\subsubsection{Control Experiments}
\begin{figure*}[t]
\begin{subfigure}[b]{0.30\textwidth}
  \centering
  \includegraphics[width=\linewidth]{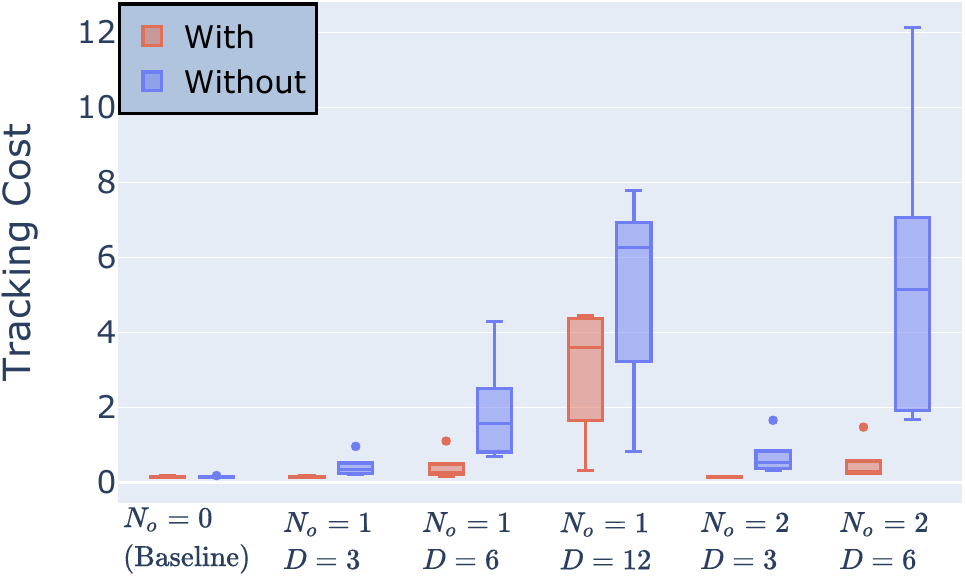}
  \caption{Trial 1}
  \label{ex0}
\end{subfigure}\hfill
\begin{subfigure}[b]{0.30\textwidth}
  \centering
  \includegraphics[width=\linewidth]{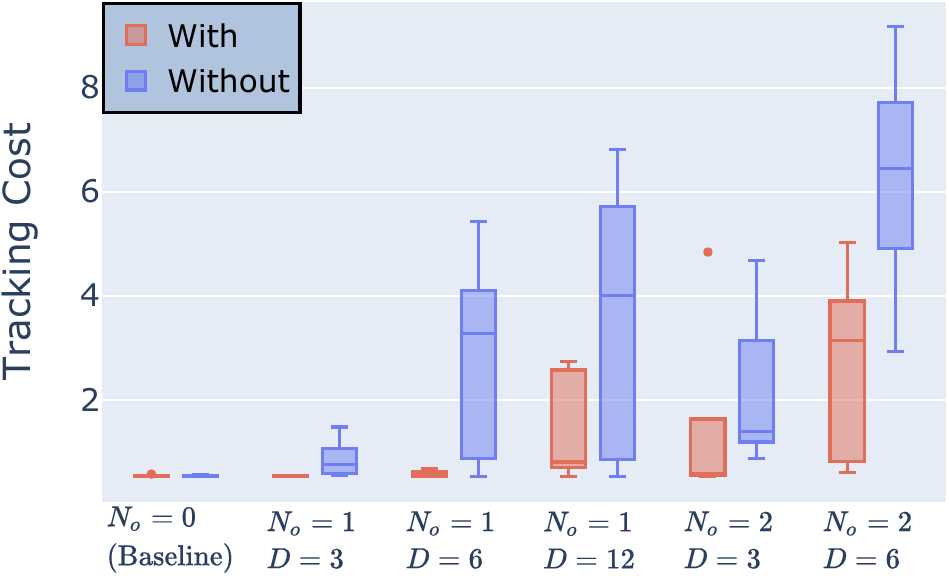}
  \caption{Trial 2}
  \label{ex1}
\end{subfigure}\hfill
\begin{subfigure}[b]{0.30\textwidth}
  \centering
  \includegraphics[width=\linewidth]{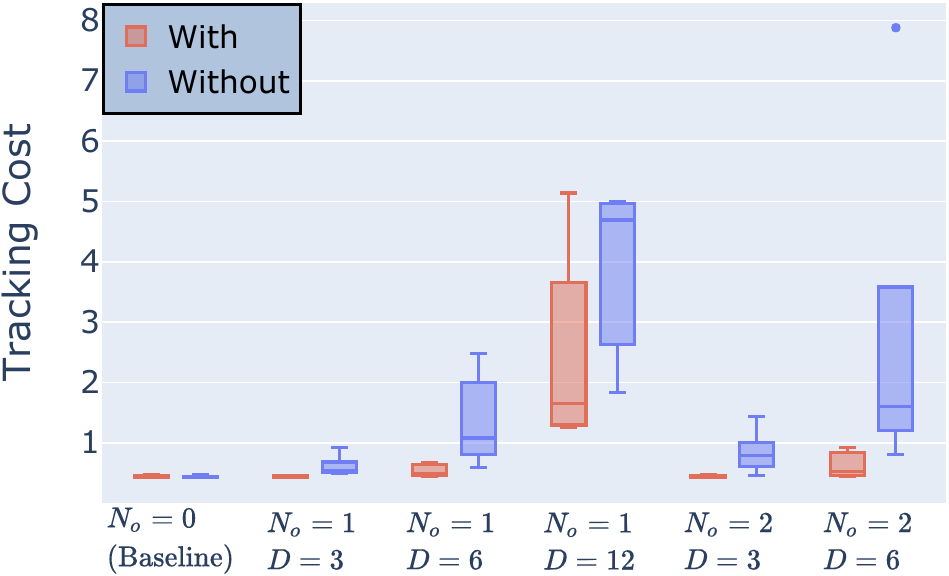}
  \caption{Trial 3}
  \label{ex2}
\end{subfigure}\\
\begin{subfigure}[b]{0.30\textwidth}
  \centering
  \includegraphics[width=\linewidth]{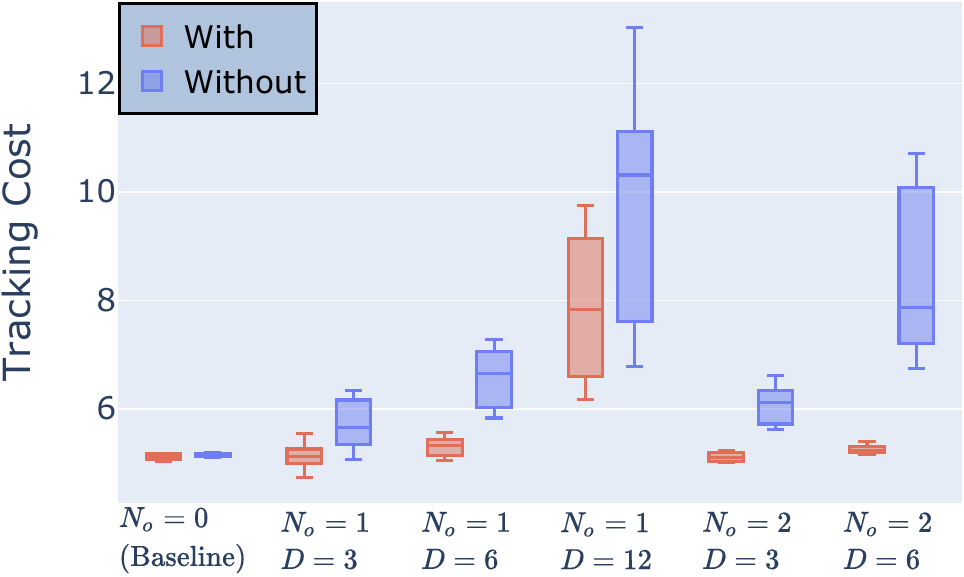}
  \caption{Trial 4}
  \label{ex3}
\end{subfigure}\hfill
\begin{subfigure}[b]{0.30\textwidth}
  \centering
  \includegraphics[width=\linewidth]{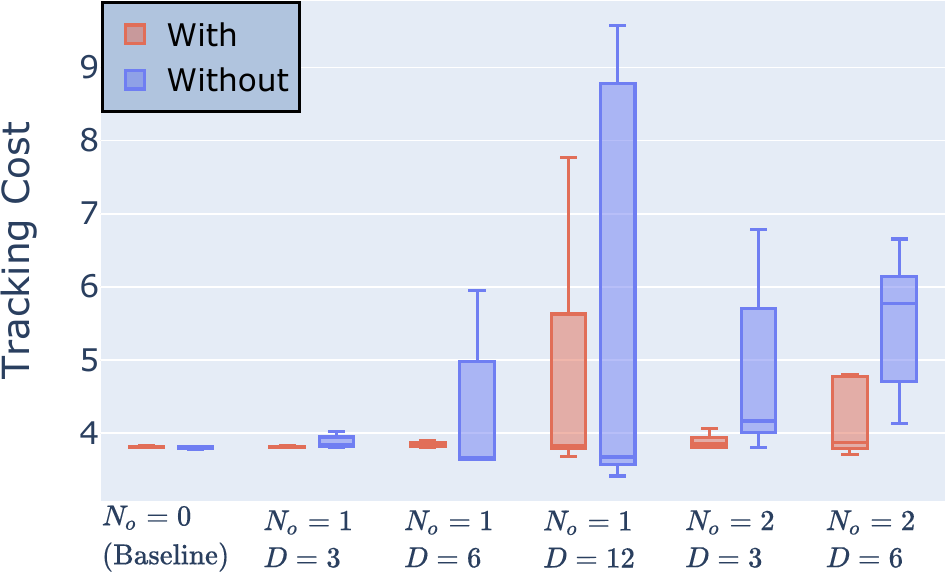}
  \caption{Trial 5}
  \label{ex4}
\end{subfigure}\hfill
\begin{subfigure}[b]{0.30\textwidth}
  \centering
  \includegraphics[width=\linewidth]{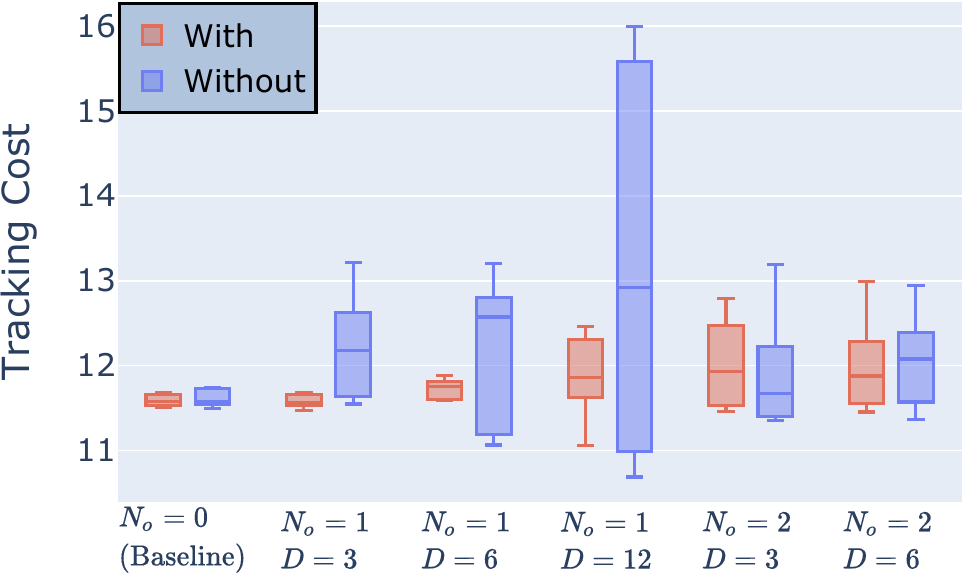}
  \caption{Trial 6}
  \label{ex5}
\end{subfigure}
\caption{Comparison of tracking costs for six trials with different initial images and the same goal image. For each trial, we tested six different occlusion settings, five times each (\emph{(a) - (f)}). We varied the number of times an occlusion was present, $N_{o}$, and the `duration' or number of images sequentially occluded, $D$. The data coloured in red with the label `With' includes the use of heteroscedastic uncertainty, while the data coloured in blue with the label `Without' does not.}
\label{fig:box_whisker}
\end{figure*}
\begin{figure*}
    \centering
	\captionsetup{aboveskip=0pt}
    \begin{subfigure}[t]{0.225\textwidth}
        \includegraphics[width=\textwidth]{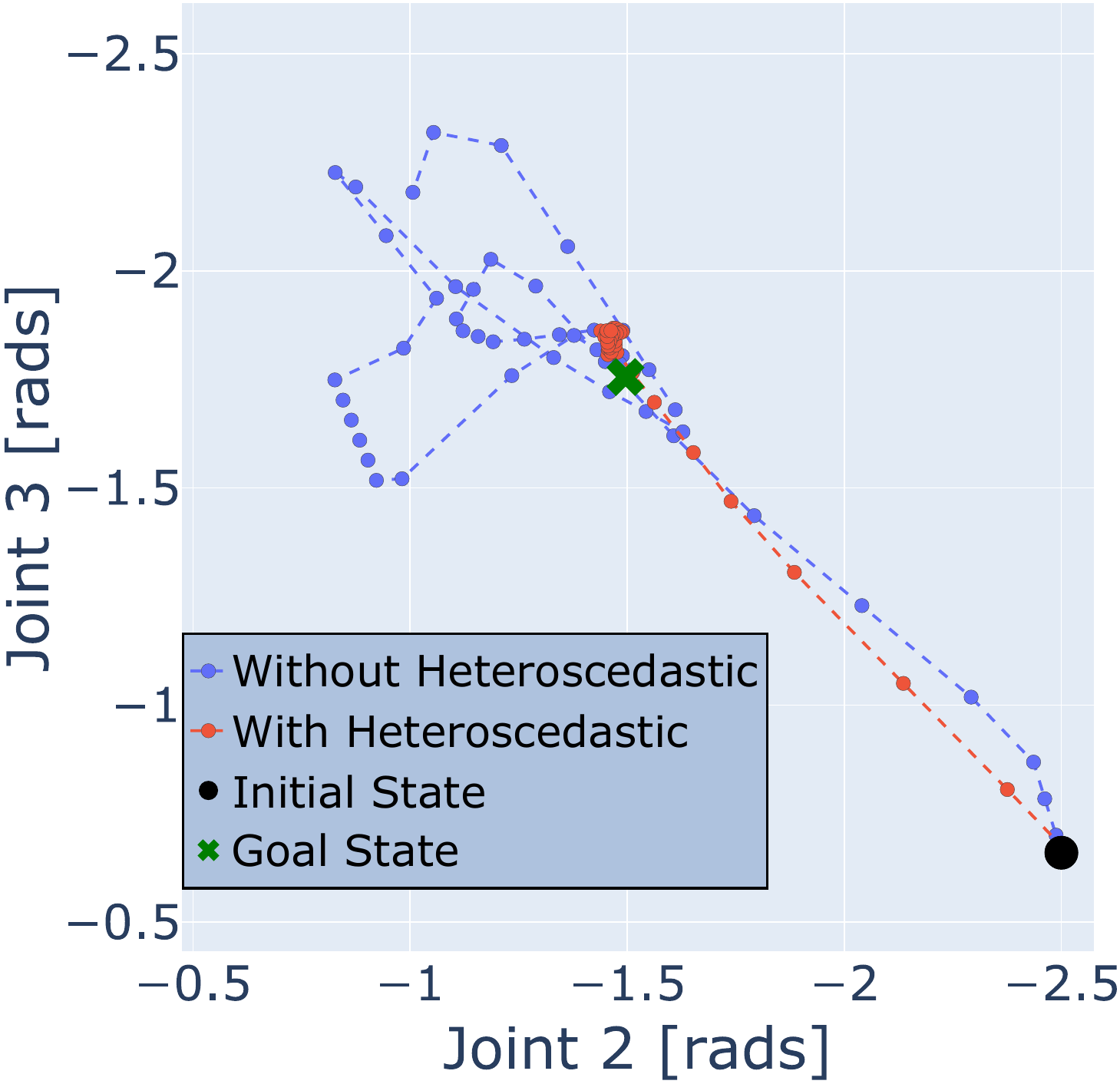}
        \caption{}
        \label{fig:traj_vis1}
        \usebox{\largestimage}
    \end{subfigure}
    ~
    \begin{subfigure}[t]{0.225\textwidth}
        \includegraphics[width=\textwidth]{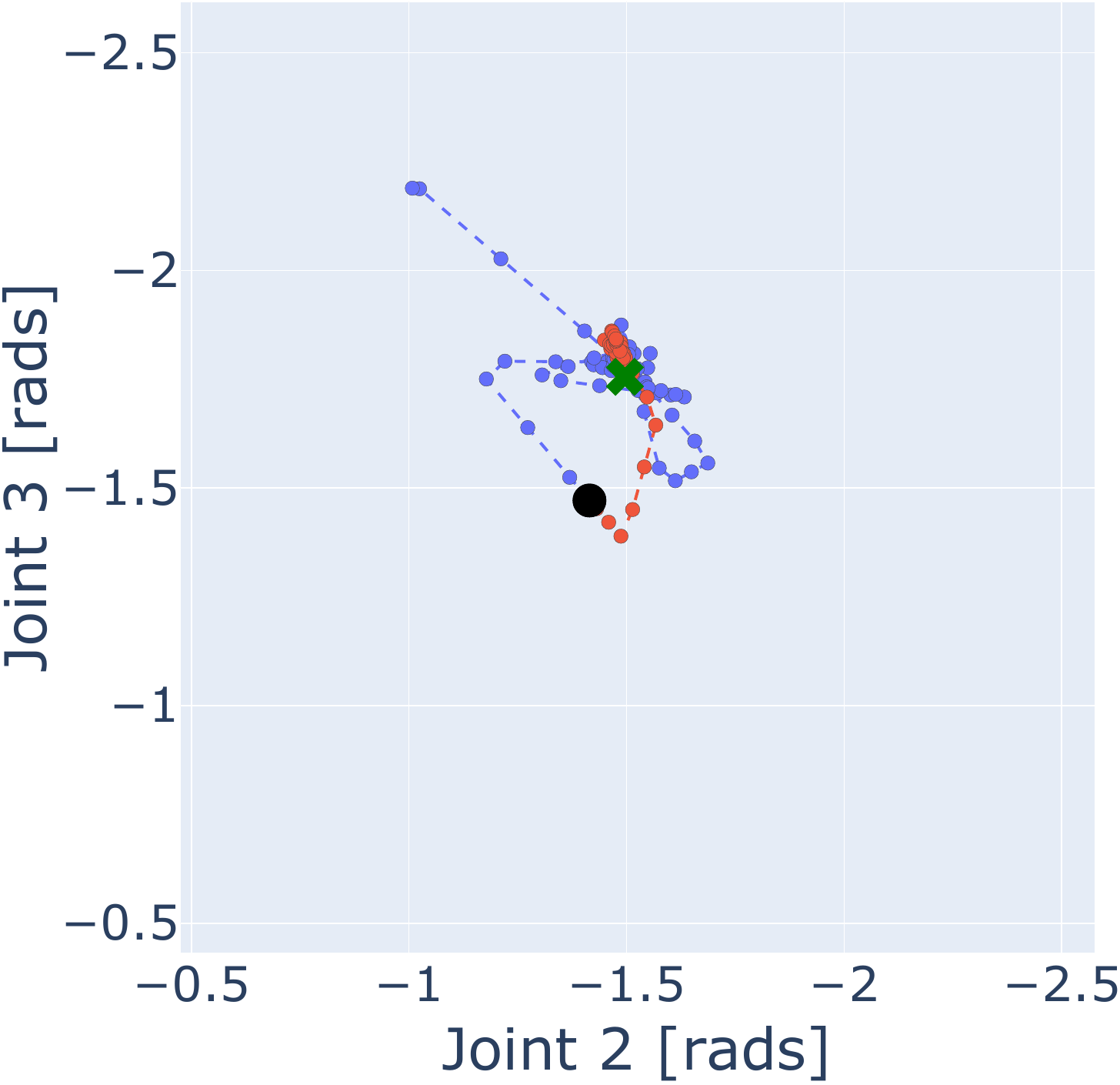}
        \caption{}
        \label{fig:traj_vis2}
    \end{subfigure}
    ~
    \begin{subfigure}[t]{0.225\textwidth}
        \includegraphics[width=\textwidth]{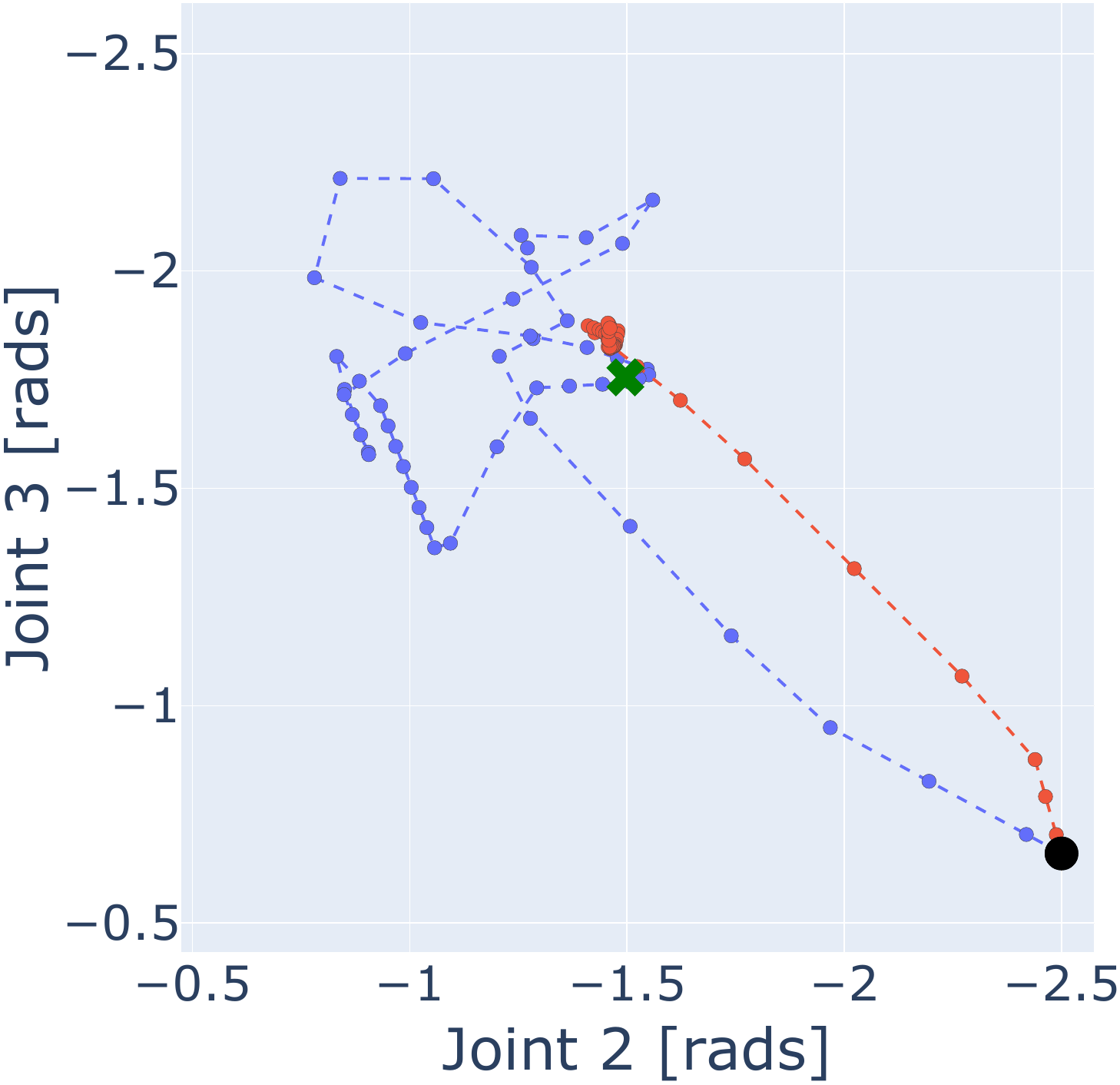}
        \caption{}
        \label{fig:traj_vis3}
    \end{subfigure}
     ~
    \begin{subfigure}[t]{0.225\textwidth}
        \includegraphics[width=\textwidth]{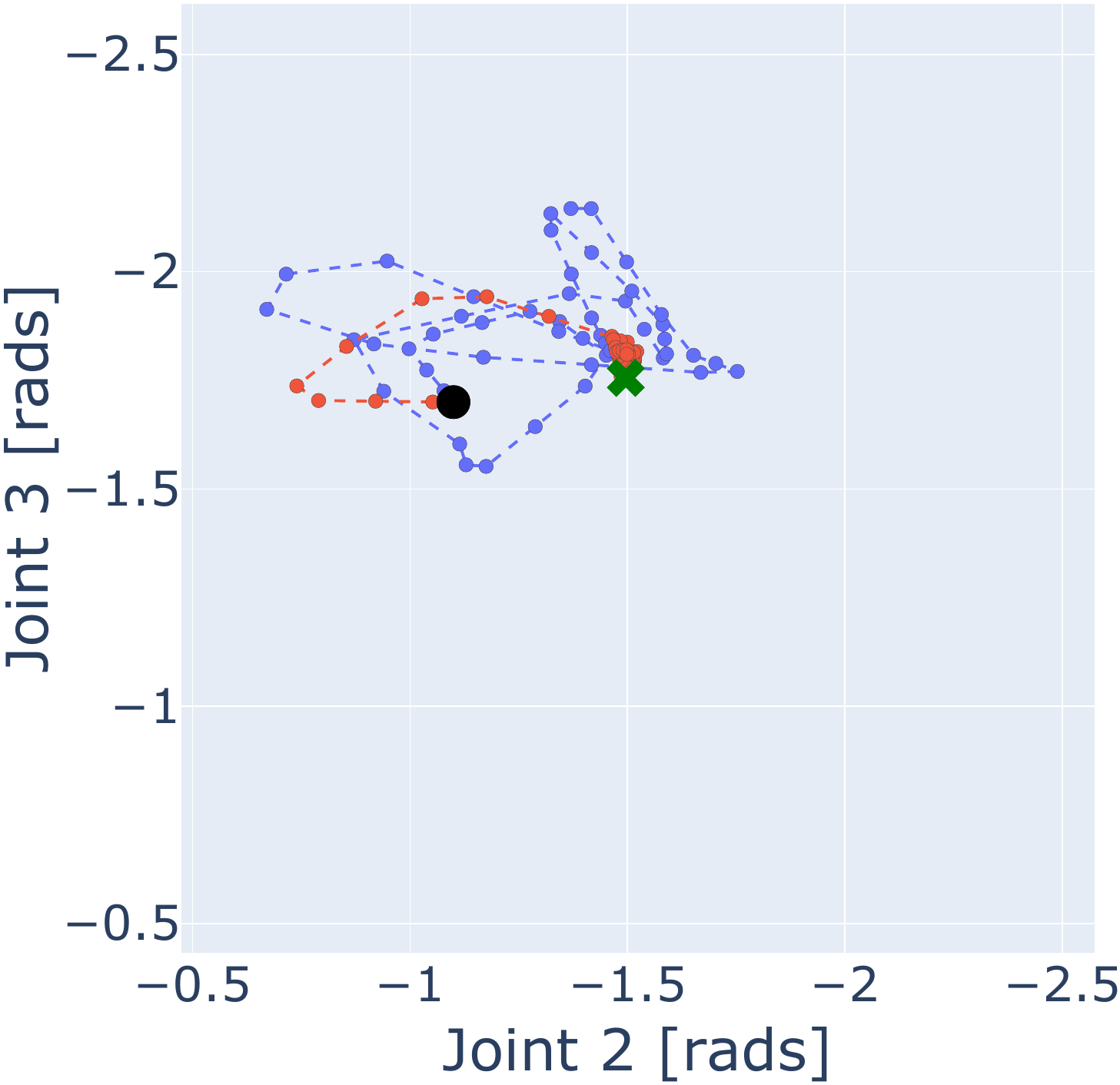}
        \caption{}
        \label{fig:traj_vis4}
    \end{subfigure}
    \caption{Visualization of four different trajectories from our control experiments, with and without heteroscedastic uncertainty weighing. The model without weighing was unable to reach and remain in the configuration shown by the goal image. The obstructions caused the manipulator to move erratically. The model with heteroscedastic uncertainty weighing was able to reach and remain at the goal.}
    \label{fig:traj_vis}
\end{figure*}

We investigated the control performance of the model by introducing a varying number of synthetic obstructions, each of which appeared for a varying duration. A summary of the results is shown in Fig. \ref{fig:box_whisker}. We used the following cumulative squared tracking error, $\sum_{k=1}^{K} \|\mathbf{q}_{\text{goal}} - \mathbf{q}_{k}\|^{2}$, where $\mathbf{q}_{\text{goal}}$ and $\mathbf{q}_{k}$ are the manipulator's joint positions for the goal image and at time step $k$, respectively.
We ran each model with the controller for $K = 24$ time steps, where each time step lasted 0.5 seconds, with the following settings for the MPC optimization: $\mathbf{Q}_{\text{mpc}} = \mathbf{I}$, $\mathbf{R}_{\text{mpc}} = \mathbf{I}$, and a prediction horizon of $T = 9$. The use of synthetic degradations allowed us to fairly perform an ablation study by comparing the performance of both models under identical conditions.
We randomly and uniformly sampled a time step or multiple time steps within the trajectory at which an obstruction was visible. We ran six different trials with different initial images and the same goal image. For each of the trials, we applied five different `classes' of occlusions, by varying the number of times an occlusion appeared, $N_{o}$, and the `duration' or quantity of images sequentially occluded, $D$. We repeated each test five times. The same random seed was used for trials with and without our heteroscedastic uncertainty weighing. In all cases, our model achieved a significantly lower tracking cost by being more robust to occlusions.
We visualize four example joint space trajectories in Fig. \ref{fig:traj_vis}, where the manipulator joint encoders provided ground truth. For these experiments, we alternated between two unobstructed images and two obstructed images. The trajectory in orange was produced by the model with heteroscedastic weighing, while the trajectory in blue was produced without this weighting. Heteroscedastic uncertainty weighing confers the ability to more consistently and accurately track the target arm pose; predictions made with homoscedastic uncertainties led to erratic control behaviour and poor performance.

\section{Conclusion}
\label{conclusion}

We have introduced a robust generative latent dynamics model capable of accurately predicting future states under difficult input degradations that (ideally) should not affect the underlying dynamical system. We presented preliminary results from experiments on two image-based tasks: a simulated visual pendulum task and a real-world manipulator visual reaching task. Our approach is a contribution towards enabling the deployment of learned generative latent dynamics models in real-world, safety-critical applications such as robotics. In this initial work, we chose simply to weight our measurement uncertainties using a reconstruction-based loss as a proxy for novel or OOD detection. However, this approach is conservative and images which are not well reconstructed may still be useful. An interesting avenue for future research would be to combine novelty detection with self-supervised uncertainty learning.

{\footnotesize
\balance
\bibliographystyle{ieeetr}
\bibliography{robotics_abbrv, paper}
}

\end{document}